\documentclass[10pt,twocolumn,letterpaper]{article}

\usepackage{wacv}
\usepackage{times}
\usepackage{epsfig}
\usepackage{graphicx}
\usepackage{amsmath}
\usepackage{amssymb}
\usepackage{pifont}

\newcommand{\xmark}{\ding{55}}%
\usepackage{array,multirow,graphicx}

\usepackage{mathtools}
\DeclarePairedDelimiter\abs{\lvert}{\rvert}%
\makeatletter
\let\oldabs\abs
\def\abs{\@ifstar{\oldabs}{\oldabs*}}
\DeclarePairedDelimiter\norm{\lVert}{\rVert}%
\let\oldnorm\norm
\def\norm{\@ifstar{\oldnorm}{\oldnorm*}}
\makeatother

\DeclareMathOperator{\EX}{\mathbb{E}}

\def\wacvPaperID{6} 

\wacvfinalcopy 

\ifwacvfinal
\def\assignedStartPage{9876} 
\fi

\usepackage[breaklinks=true,bookmarks=false]{hyperref}

\ifwacvfinal
\else
\fi

\begin{document}

\title{Same Same But DifferNet: \\Semi-Supervised Defect Detection with Normalizing Flows}

\author{Marco Rudolph \hspace{2cm} Bastian Wandt \hspace{2cm} Bodo Rosenhahn
\\Leibniz University Hanover\\
{\tt\small \{rudolph, wandt, rosenhahn\}@tnt.uni-hannover.de}}

\maketitle

\begin{abstract}
    The detection of manufacturing errors is crucial in fabrication processes to ensure product quality and safety standards.
    Since many defects occur very rarely and their characteristics are mostly unknown a priori, their detection is still an open research question.
    To this end, we propose \textnormal{DifferNet}: It leverages the descriptiveness of features extracted by convolutional neural networks to estimate their density using normalizing flows.
    Normalizing flows are well-suited to deal with low dimensional data distributions.
    However, they struggle with the high dimensionality of images.
    Therefore, we employ a multi-scale feature extractor which enables the normalizing flow to assign meaningful likelihoods to the images.
    Based on these likelihoods we develop a scoring function that indicates defects.
    Moreover, propagating the score back to the image enables pixel-wise localization.
    To achieve a high robustness and performance we exploit multiple transformations in training and evaluation.
    In contrast to most other methods, ours does not require a large number of training samples and performs well with as low as 16 images.
    We demonstrate the superior performance over existing approaches on the challenging and newly proposed MVTec AD \cite{mvtec} and Magnetic Tile Defects \cite{magnets} datasets.

\end{abstract}

\begin{figure}
\centering
  \includegraphics[width=0.45\textwidth]{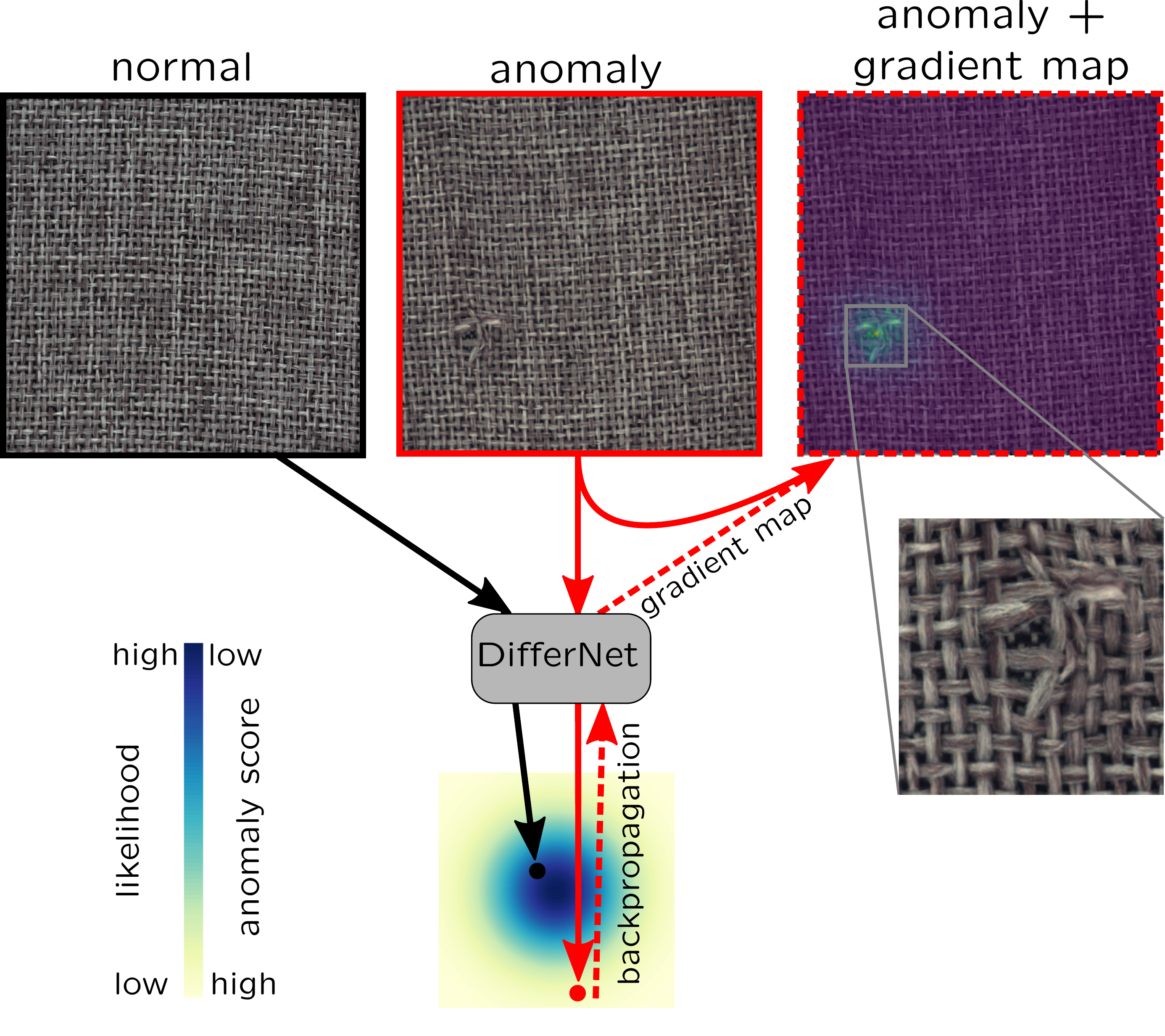} 
 \caption{DifferNet assigns likelihoods to inputs which makes it usable to detect defects. In contrast to the top left image without any defect, the top middle image is assigned to a high anomaly score due to the defect (see the enlarged patch on the right side). Additionally, DifferNet identifies the defective region by backpropagating the likelihood loss up to the input which gives a gradient map (top right image). This allows for a detailed analysis of defect position and shape.}
\label{fig:teaser}
\end{figure}
\section{Introduction}
In industrial manufacturing processes the quality of the products is constantly monitored and improved.
Hence, small defects during fabrication need to be detected reliably.
However, manufacturers do not know in advance which types of defects will occur and most of them appear so infrequently that no defective examples are available.
Even if some defect types are known, new types can still occur any time due to unforeseeable events during manufacturing.
Consequently, reliable defect detection cannot be done with supervised machine learning approaches.
We propose a solution for semi-supervised defect detection where only positive examples and no defective examples are present during training.
This is also known as anomaly detection.

In general, anomaly detection describes the problem of determining whether a data sample differs from a set of given \textit{normal} data.
There are various approaches for general anomaly detection on images, summarized in Section~\ref{related}.
Defect detection is a specific sub-problem where visually similar normal samples and only slightly different anomalous samples are present.
While traditional anomaly detection methods are well-suited to data with high intra-class variance, they are not able to capture subtle differences.
We tackle this problem by employing an accurate density estimator on image features extracted by a convolutional neural network.
The feature distribution of normal samples is captured by utilizing the latent space of a normalizing flow \cite{nf}.
Unlike other generative models such as variational autoencoders \cite{vae} or GANs \cite{gan}, there exists a bijective mapping between feature space and latent space in which each vector is assigned to a likelihood.
This enables DifferNet to calculate a likelihood for each image.
From this likelihood we derive a scoring function to decide if an image contains an anomaly.
Figure~\ref{fig:teaser} visualizes the core idea behind our method.
The most common samples are assigned to a high likelihood whereas uncommon images are assigned to a lower likelihood.
Since defects are not present during training, they are mapped to a low likelihood.
We further improve the descriptiveness of the feature extractor by using multi-scale inputs.
To derive a meaningful scoring function, we include likelihoods of several transformations of the image.
Thereby DifferNet gains flexibility and robustness to detect various types of defects.
We show that our method considers even small changes in data while other approaches struggle to detect them.
Moreover, due to the efficiency of the feature extractor and the proposed image transformations, our approach even outperforms the state of the art when trained with a low number of training samples.
Besides defect detection, DifferNet's architecture allows for localization of defects by having expressive gradients of input pixels regarding the scoring functions.
A high magnitude of the gradient in an image region signals an area with anomalous features which helps to identify the defect.

Our work comprises the following contributions:
\begin{itemize}
    \item Detection of anomalies via the usage of likelihoods provided by a normalizing flow on multi-scale image features with multi-transform evaluation.
    \item Anomaly localization without training labels, the necessity of any pixel-wise optimization and sub-image detection.
    \item Applicability on small training sets. Even with a low number of training examples our approach achieves competitive results.
    \item State-of-the-art detection performance on MVTec AD and Magnetic Tile Defects. 
    \item Code is available on GitHub \footnote{\url{https://github.com/marco-rudolph/differnet}}.
\end{itemize}

\section{Related Work}
\label{related}
\subsection{Anomaly Detection}
Existing methods for anomaly detection can be roughly divided into approaches based on generative models and pretrained networks.
The most relevant methods to our work are briefly presented in the following subsections.
Note that we focus on works that deal with image anomaly detection rather than anomaly localization to keep the focus on our main problem.

\subsubsection{Detection with Pretrained Networks}
There are several works which use the feature space of a pretrained network to detect anomalies.
In most cases simple traditional machine learning approaches are used to obtain an anomaly score.
Andrews \etal \cite{andrews} apply a One-Class-SVM on VGG \cite{VGG} features of the training images.
Nazare \etal \cite{nazare} evaluated different normalization techniques to have a 1-Nearest-Neighbor-classifier on a PCA-reduced representation of the feature space.
Localization of the anomalies is achieved by evaluating the method on many overlapping patches. However, this is very costly.
Sabokrou \etal \cite{sabokrou} models the anomaly-free feature distribution as an unimodal Gaussian distribution. Therefore, they cannot capture multimodal distributions as opposed to our method.

These techniques only work for particular classes in defect detection.
In contrast to our proposed DifferNet, existing approaches do not appear to be robust and powerful enough for defect detection.
None of the above techniques take advantage of the flexibility of another neural network on top of the pretrained model.
Another benefit of our method compared to these approaches is being able to compute gradients \wrt the inputs which can be utilized to compute anomaly maps.

\begin{figure*}[ht!]
\centering
  \includegraphics[width=0.96\textwidth]{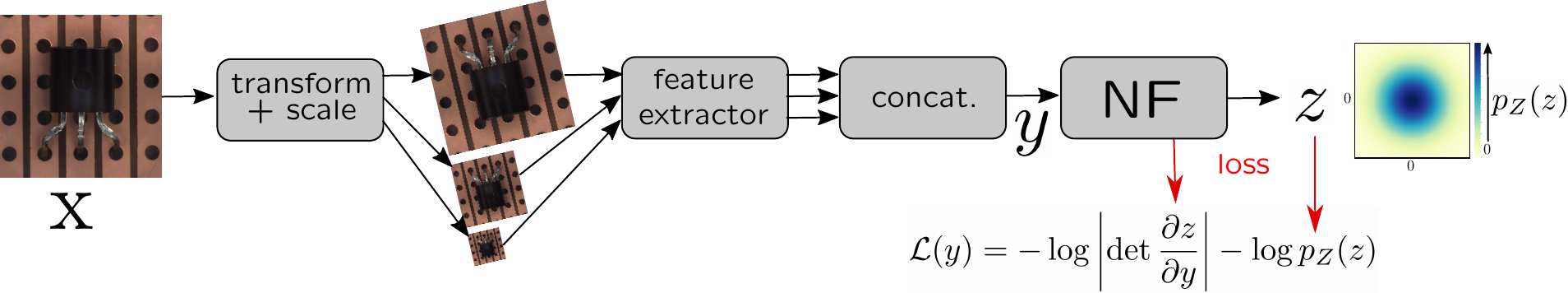} 
 \caption{Overview of our pipeline: Multiple scales of a transformed input image are fed into a feature extractor. The distribution of its concatenated outputs is captured by transforming it via a normalizing flow (NF) into a normal distribution by maximum likelihood training.}
\label{fig:overview}
\end{figure*}

\subsubsection{Generative Models}
Generative models, such as autoencoders \cite{ae_lecun,vae,sae} and GANs \cite{gan}, are able to generate samples from the manifold of the training data.
Anomaly detection approaches using these models are based on the idea that the anomalies cannot be generated since they do not exist in the training set.

Autoencoder-based approaches try to detect anomalies by comparing the output of an autoencoder to its input.
Thus, a high reconstruction error should indicate an anomalous region.
Bergmann \etal \cite{ae_ssim} proposes SSIM-loss to make the reconstruction error dependent on visual similarity.
In many cases autoencoder-based methods fail because they generalize too strongly, \ie anomalies can be reconstructed as good as normal samples.
Gong \etal \cite{memae} tackle the generalization problem by employing memory modules which can be seen as a discretized latent space.
Zhai \etal \cite{dsebm} connect regularized autoencoders with energy-based models to model the data distribution and classify samples with high energy as an anomaly.

GAN-based approaches assume that only positive samples can be generated.
Schlegl \etal \cite{anogan} propose a two-stage training method: The GAN is learned first and an encoder is optimized as an inverse generator.
Using the generator as decoder enables the calculation of a reconstruction loss alongside the difference in discriminator features of original and reconstructed image to obtain an anomaly score.
Akcay \etal \cite{ganomaly} make use of adversarial training by letting an autoencoder directly act as generating part of the GAN.
This enforces the property of the decoder to only generate normal-like samples which can be measured by the difference between the embedding of the original and the reconstructed data.

We argue that generative models are appropriate for a wide range of defect detection scenarios since they strongly depend on the anomaly type.
For example, the size and structure of the defective area heavily influence the anomaly score.
If the region of interest shows high frequency structures, they cannot be represented accurately.
Often, other instance-specific structures influence the reconstruction error more than the anomaly.
In contrast, we show that DifferNet handles significantly more and various defect types. 
Additionally, our method does not rely on a large number of training samples compared to generative models.

\subsection{Normalizing Flows}
Normalizing Flows (NF) \cite{nf} are neural networks that are able to learn transformations between data distributions and well-defined densities.
Their special property is that their mapping is bijective and they can be evaluated in both directions.
First, they assign likelihoods to a given sample.
Second, data is generated by sampling from the modeled distribution.
The bijectivity is ensured by stacking layers of affine transforms which are fixed or autoregressive.
A common class of such autoregressive flows is \textit{MADE} (Germain \etal \cite{germain}) which makes use of the Bayesian chain rule to decompose the density.
These models can learn distributions of large datasets containing of mostly small images.
In contrast, we capture the distribution of a comparably small number of images at a high resolution.
Autoregressive flows compute likelihoods fast, but are slow at sampling.
Inverse autoregressive flows, proposed by Kingma \etal \cite{kingma}, show the exact opposite behavior.
Real-NVP \cite{realnvp} can be seen as a special inverse autoregressive flow which is simplified such that both forward and backward pass can be processed quickly.
Similar to Ardizzone \etal proposed for \textit{Invertible Neural Networks} \cite{cinn}, we integrate a parametrized clamping mechanism to the affine transformation of the Real-NVP-layers to obtain a more stable training process.
Details to the so-called coupling layers and the clamping mechanism of Real-NVP are explained in Section~\ref{normflow}.

The property of normalizing flows as an adequate estimator of probability densities to detect anomalies has not raised much attention yet, although some works present promising results using Real-NVP and MADE \cite{nf_deep, nf_time_series, nf_trajectory}.
However, none of the works deal with visual data.

\section{Method}
\label{method}
Figure~\ref{fig:overview} shows an overview of our pipeline.
Our method is based on density estimation of image features $y \in Y$ from the anomaly-free training images $x \in X$.
Let $f_{\mathrm{ex}}: X \xrightarrow{} Y$ be the mapping of a pretrained feature extractor which is not further optimized. 
The estimation of $p_Y(y)$, provided by $f_{\mathrm{ex}}(x)$, is achieved by mapping from $Y$ to a latent space $Z$ -- with a well-defined distribution $p_Z(z)$ -- by applying a normalizing flow $f_{\mathrm{NF}}: Y \xrightarrow{} Z$.
Likelihoods for image samples are directly calculated from $p_Z(z)$.
Features of anomalous samples should be out of distribution and hence have lower likelihoods than normal images. 
Likelihoods of multiple transforms on the image are maximized in training and used in inference for a robust prediction of the anomaly score.
To capture structures at different scales and thus having a more descriptive representation in $y$, we further define $f_{\mathrm{ex}}$ as the concatenation of features at 3 scales.

\begin{figure*}
\centering
  \includegraphics[width=1\textwidth]{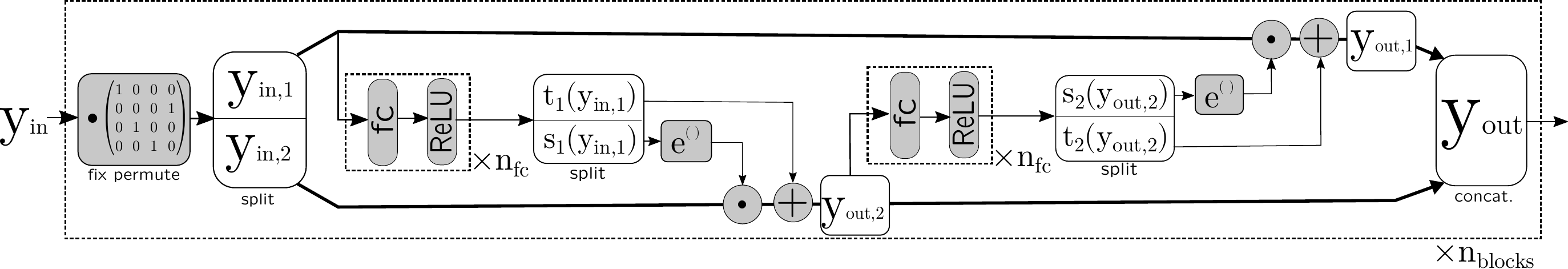} 
 \caption{Architecture of one block inside the normalizing flow: After a fixed random permutation, the input is split into two parts that regress scale and shift parameters to transform their respective counterpart. Symbols $\odot$ and $\oplus$ denote element-wise multiplication and addition, respectively. Numerical operations are symbolized by grey blocks. White blocks contain variable names.}
\label{fig:inn_block}
\end{figure*}
\subsection{Normalizing Flow}
\label{normflow}
The normalizing flow acts as a bijector between feature space $Y$ and latent space $Z$ by using affine transformations.
Likelihoods are computed from $Z$ according to the modelled distribution.
We model $z \sim \mathcal{N}(0,\,I)$ in our latent space which gives us a well-defined density $p_Z(z)$ for $z$.
Following from bijectivity, for every feature input there is a unique $z$ of the same dimension and vice versa.

\subsubsection{Architecture}
We use an architecture of coupling layers as proposed in Real-NVP \cite{realnvp}.
The detailed structure of one block is shown in Figure~\ref{fig:inn_block}. The design of $f_{\mathrm{NF}}$ is a chain of such blocks, consisting of permutations followed by scale and shift operations.

To apply the scale and shift operations, the input $y_{\mathrm{in}}$ is split into $y_{\mathrm{in}, 1}$ and $y_{\mathrm{in}, 2}$ that manipulate each other by regressing multiplicative and additive components in subnetworks $s$ and $t$; these manipulations are applied to their respective counterpart successively.
The scale and shift operations are described by
\begin{equation}
 \begin{aligned}
y_{\mathrm{\mathrm{out}}, 2} = y_{\mathrm{in}, 2} \odot e^{s_1(y_{\mathrm{in}, 1})} + t_1(y_{\mathrm{in}, 1}) \\
y_{\mathrm{\mathrm{out}}, 1} = y_{\mathrm{in}, 1} \odot e^{s_2(y_{\mathrm{\mathrm{out}}, 2})} + t_2(y_{\mathrm{\mathrm{out}}, 2}),
\end{aligned}
\end{equation}
with $\odot$ as the element-wise product. 
Using an exponential function before scaling preserves the affinity property by ensuring non-zero coefficients.
The internal functions $s$ and $t$ can be any differentiable function, which in our case is implemented as a fully-connected network that regresses both components by splitting the output (see Figure~\ref{fig:inn_block}).
Similar to Ardizzone \etal \cite{cinn}, we apply soft-clamping to the values of $s$ to preserve model stability which is crucial in our case for better convergence. 
This is achieved by using the activation
\begin{equation}
\sigma_{\alpha}(h) = \frac{2\alpha}{\pi}\arctan{\frac{h}{\alpha}}
\end{equation}
as the last layer of $s$.
This prevents large scaling components by restricting $s(y)$ to the interval $(-\alpha, \alpha)$.

Each block first performs a predefined random permutation on the features to allow each dimension to affect all other dimensions at some point.
The output of one coupling block is given by the concatenation of $y_{\mathrm{out}, 1}$ and $y_{\mathrm{out}, 2}$.

\subsubsection{Training}
\label{Training}
The goal during training is to find parameters for $f_{\mathrm{NF}}$ that maximize likelihoods for extracted features $y \in Y$ which are quantifiable in the latent space $Z$. With the mapping $z=f_{\mathrm{NF}}(y)$ and according to the change-of-variables formula Eq.~\ref{eqn:change_of_variables}, we describe this problem as maximizing
\begin{equation}
\label{eqn:change_of_variables}
    p_Y(y) = p_Z(z) \abs{
    \det{
    \frac{\partial z}
        {\partial y}
    }}
    .
\end{equation}
This is equivalent to maximizing the log-likelihood, which is more convenient here, since the terms simplify when inserting the density function of a standard normal distribution as $p_Z(z)$. We use the negative log-likelihood loss $\mathcal{L}(y)$ to obtain a minimization problem:
\begin{equation}
 \begin{aligned}
    \log{p_Y(y)} = \log{p_Z(z)}  + \log{\abs{
    \det{
    \frac{\partial z}
        {\partial y}
    }}}\\
    \mathcal{L}(y) = \frac{\norm{z}_2^2}{2}    - \log{\abs{
    \det{
    \frac{\partial z}
        {\partial y}
    }}}
    .
 \end{aligned}
 \label{formula:loglikelihood}
\end{equation}
Intuitively, we want $f_{\mathrm{NF}}$ to map all $y$ as close as possible to $z=0$ while penalizing trivial solutions with scaling coefficients close to zero\footnote{The exponentiation inhibits the coefficients from being zero.}.
The latter in ensured by the negative log determinant of the Jacobian $\frac{\partial z} {\partial y}$ in $\mathcal{L}(y)$.
In our case the log determinant of the Jacobian is the sum of scaling coefficients before exponentiation. 

During Training, $\mathcal{L}(y)$ is optimized for features $y$ of different transformations of an input image for a fixed epoch length. Section~\ref{implementation} describes the training in more detail.

\subsection{Scoring Function}
We use the calculated likelihoods as a criterion to classify a sample as anomalous or normal.
To get a robust anomaly score $\tau(x)$, we average the negative log-likelihoods using multiple transformations $T_i(x) \in \mathcal{T}$ of an image $x$:
\begin{equation}
    \tau(x) = \EX_{T_i \in \mathcal{T} } {[-\log{p_Z(f_{\mathrm{NF}}(f_{\mathrm{ex}}(T_i(x))))}]}
    .
\end{equation}

As $\mathcal{T}$ we choose rotations and manipulations of brightness and contrast.
An image is classified as anomalous if the anomaly score $\tau(x)$ is above the threshold value $\theta$. 
Thus, the decision can be expressed as
\begin{equation}
    \mathcal{A}(x) = 
      \begin{cases}
        1 & \text{for } \tau(x) \geq \theta \\
        0 & \text{for } \tau(x) < \theta 
      \end{cases}
      ,
\end{equation}
where $\mathcal{A}(x)=1$ indicates an anomaly.
In Section~\ref{experiments} $\theta$ is varied to calculate the \textit{Receiver Operating Characteristic} (ROC).

\subsection{Localization}
\label{loc_method}
In contrast to several other approaches, ours is not optimized for localizing the defects on the image.
Nevertheless, our method localizes areas where anomalous features occur.
Our method allows for propagating the negative log-likelihood $\mathcal{L}$ back to the input image $x$.
The gradient $\nabla x_c$ of each input channel $x_c$ is a value indicating how much the pixels influence the error which relates to an anomaly.
For better visibility we blur these gradients with a Gaussian kernel $G$ and sum the absolute values over the channels $C$ according to
\begin{equation}
    g_x = \sum_{c \in C}{\abs{G * \nabla x_c}}
    ,
\end{equation}
with $*$ as 2D convolution and $|\cdot|$ as the element-wise absolute value, which results in the gradient map $g_x$.
Averaging the maps of multiple rotations of one single image -- after rotating back the obtained maps -- gives a robust localization.

\section{Experiments}
\label{experiments}

\begin{figure}
\centering
  \includegraphics[width=0.45\textwidth]{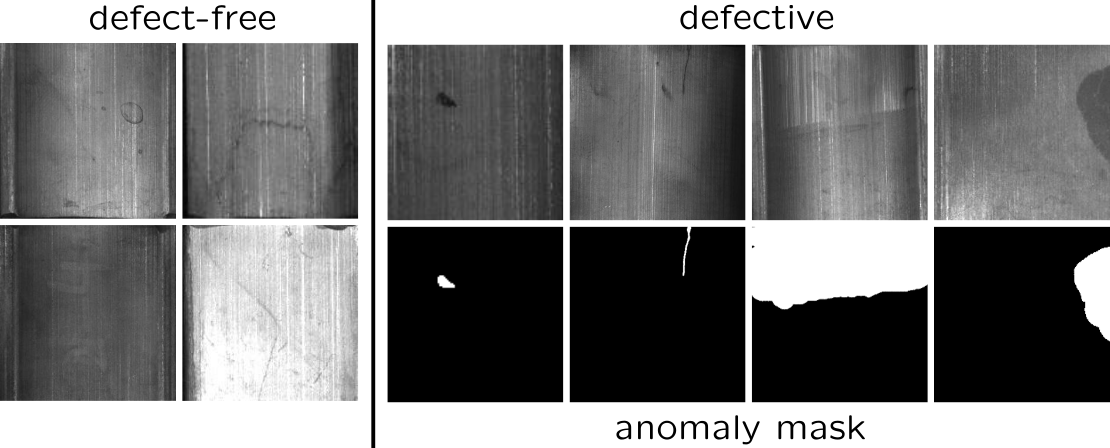} 
 \caption{Samples of defect-free and defective images from Magnetic Tile Defects \cite{magnets}.}
\label{fig:magnet_examples}
\end{figure}

\subsection{Datasets}
In this paper, we evaluate our approach on real-world defect detection problems. 
We use the challenging and newly proposed datasets MVTec AD \cite{mvtec} and Magnetic Tile Defects (MTD) \cite{magnets}.
The difficulty in these datasets lies in the similarity of anomalies and normal examples.

To the best of our knowledge, MVTec AD is the only publicly available multi-object and multi-defect anomaly dataset.
It contains 5354 high-resolution color images of 10 object and 5 texture categories.
The number of training samples per category ranges from 60 to 320, which is challenging for the estimation of the distribution of normal samples.
Several defect types per category, such as little cracks, deformations, discolorizations and scratches are occurring in the test set.
Some of which are shown in Figure~\ref{fig:localization}.
In total, the dataset includes 70 defect types.
The anomalies differ in their size, shape and structure and thus cover several scenarios in industrial defect detection.

Magnetic Tile Defects \cite{magnets} comprises grayscale images of magnetic tiles under different illuminations with and without defects.
Such tiles should provide a constant magnetic potential in engines.
We split the 952 defect-free images randomly into a train and a test set, where the test set contains 20\% of the data.
All 392 defect images are used for testing.
These show frayed or uneven areas, cracks, breaks and blowholes as anomalies, as shown in Figure~\ref{fig:magnet_examples}.
A lot of defect-free images contain variations that are similar to anomalies.

\begin{table*}
\begin{center}
\begin{tabular}{c|l|c|c|c|c|c|c|c|c|}
\cline{2-9}
& Category & GeoTrans & GANomaly & DSEBM & OCSVM & 1-NN & \textbf{DifferNet} & DifferNet\\
 & & \cite{geotrans} &\cite{ganomaly}  & \cite{dsebm} & \cite{andrews}& \cite{nazare}& \textbf{(ours)} & (16 shots)\\
\cline{2-9}
& Grid       & 61.9 & 70.8 & \underline{71.7} & 41.0 & 55.7 & \textbf{84.0} & 65.8\\
& Leather    & 84.1 & 84.2 & 41.6 & 88.0 & 90.3 & \textbf{97.1} & \underline{92.9}\\
& Tile       & 41.7 & 79.4 & 69.0 & 87.6 & 96.9 & \textbf{99.4} & \underline{98.9}\\
\rotatebox[origin=c]{90}{\parbox[c]{0cm}{Textures}}& Carpet     & 43.7 & 69.9 & 41.3 & 62.7 & \underline{81.1} & \textbf{92.9} & 77.0\\
& Wood       & 61.1 & 83.4 & 95.2 & 95.3 & 93.4 & \textbf{99.8} & \underline{99.2}\\
\cline{2-9}
& Bottle     & 74.4 & 89.2 & 81.8 & \textbf{99.0}& 98.7 & \textbf{99,0}  & 98.5\\
& Capsule    & 67.0 & \underline{73.2} & 59.4 & 54.4 & 71.1 & \textbf{86.9} & 61.4\\
& Pill       & 63.0 & 74.3 & 80.6 & 72.9 & \underline{83.7} & \textbf{88.8} & 65.1 \\
& Transistor & \underline{86.9} & 79.2 & 74.1 & 56.7 & 75.6 & \textbf{91.1} & 76.6\\
& Zipper     & 82.0 & 74.5 & 58.4 & 51.7 & \underline{88.6} & \textbf{95.1} & 88.3 \\
\rotatebox[origin=c]{90}{\parbox[c]{0cm}{Objects}}& Cable      & 78.3 & 75.7 & 68.5 & 80.3 & \underline{88.5} & \textbf{95.9} & 86.4 \\
& Hazelnut   & 35.9 & 78.5 & 76.2 & 91.1 & \underline{97.9} & \textbf{99.3} & 97.3\\
& Metal Nut  & \underline{81.3} & 70.0 & 67.9 & 61.1 & 76.7 & \textbf{96.1} & 77.7\\
& Screw      & 50.0 & 74.6 & \textbf{99.9} & 74.7 & 67.0 & \underline{96.3} & 75.9\\
& Toothbrush & \underline{97.2} & 65.3 & 78.1 & 61.9 & 91.9 & \textbf{98.6} & 92.3\\
\cline{2-9}
& \textbf{Average} & 67.2 & 76.2 & 70.9 & 71.9 & 83.9 &\textbf{94.9} & \underline{87.3}\\
\cline{2-9}
\end{tabular}
\end{center}
\caption{Area under ROC in \% for detected anomalies of all categories of MVTec AD \cite{mvtec} grouped into textures and objects. Best results are in bold, second best underlined. OCSVM and 1-NN are calculated on the same feature extractor outputs our NF is trained on. \textit{16 shots} denotes a model trained on only $16$ images.}
\label{table:mvtec}
\end{table*}

\subsection{Implementation Details}
\label{implementation}
For all experiments, we use the convolutional part of AlexNet \cite{alexnet} as the feature extractor and apply global average pooling on each feature map at each scale.
We tested more complex topologies, for instance ResNet \cite{resnet} and VGG \cite{VGG}, but did not observe better performance.
We prefer the smaller AlexNet since it is sufficient for our purpose.
The feature extractor is pretrained on ImageNET and remains fixed during training.
We use features at $3$ scales with input image sizes of $448\times448$, $224\times224$ and $112\times112$ pixels - resulting in $3 \cdot 256 = 768$ features.
The normalizing flow consists of 8 coupling blocks with fully connected networks as internal functions $s$ and $t$.
These include 3 hidden dense layers with a size of 2048 neurons and ReLU activations.
We set the mentioned clamping parameter $\alpha=3$.
For training, we use the Adam Optimizer \cite{adam} with the author-suggested $\beta$- parameters and a learning rate of $2 \cdot 10^{-4}$.
As transformations $\mathcal{T}$, random rotations, which are uniformly distributed in the interval $[0, 2\pi]$, are applied.
In addition, we manipulated the contrast and brightness of the magnetic tiles with an uniformly distributed random factor in the interval $[0.85, 1.15]$.
In each case of training and inference the same transformations are applied.
We train our models for 192 epochs with a batch size of 96.

\begin{figure}
\centering
  \includegraphics[width=0.43\textwidth]{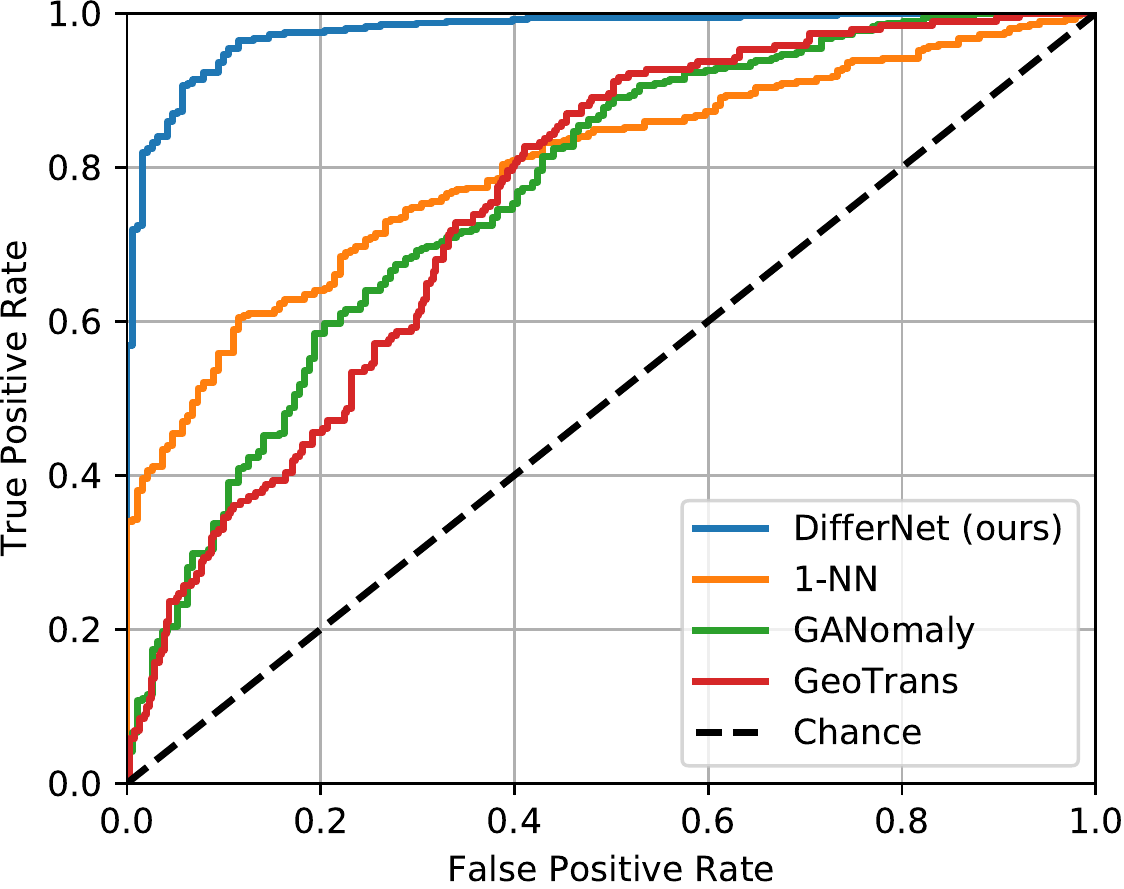} 
 \caption{ROC-Curve for different methods for detecting defects in MTD. DifferNet is significantly more accurate in detecting the defects compared to other approaches. Best viewed in color.}
\label{fig:magnetroc}
\end{figure}

\begin{figure}
\centering
  \includegraphics[width=0.43\textwidth]{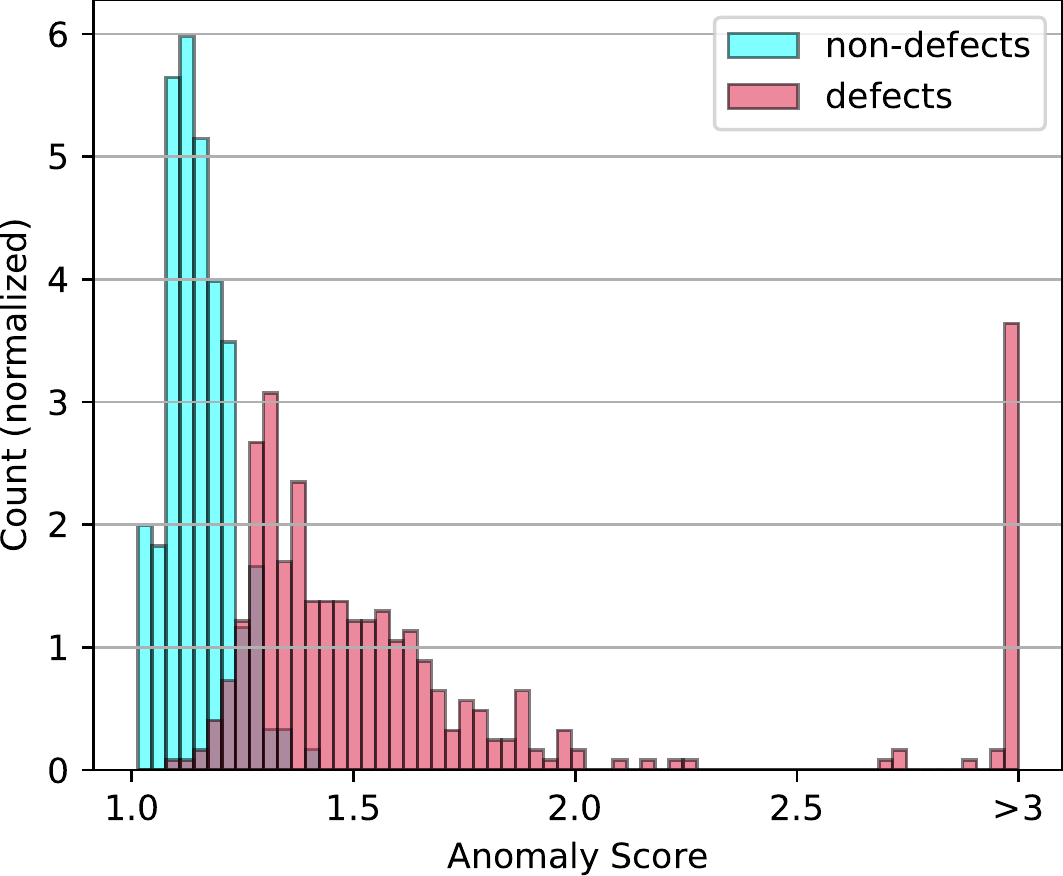} 
 \caption{Normalized histogram of DifferNet's anomaly scores for the test images of MTD. As can be seen, the score is a reliable indicator for defects except for a narrow range of some borderline cases. Note that the rightmost bar summarizes all scores above $3$.}
\label{fig:scores}
\end{figure}
\subsection{Detection}
For reporting the performance of our method regarding the detection of anomalies, we follow \cite{ganomaly} and compute the Area Under Receiver Operator Characteristics (AUROC) depending on the scoring function which we obtained as described in Section~\ref{method}.
It measures the area under the true positive rate as a function of the false positive rate.
The AUROC metric is not sensitive to any threshold or the percentage of anomalies in the test set.
Besides other anomaly detection methods, we compare our method to the baselines one-class SVM (OCSVM) \cite{andrews} and the distance to the nearest neighbor (1-NN) after PCA reduction to 64 dimensions and z-score normalization \cite{nazare}.
Note that both methods OCSVM and 1-NN are adapted to our setting:
We used every technique of our pipeline (see Figure~\ref{fig:overview}) but replaced the normalizing flow with them.
We evaluate several transformations and used the mean respective score. 
Apart from these approaches and some state-of-the-art models, we compare our method with GeoTrans \cite{geotrans} which cannot be assigned to generative and pretrained methods described in Section~\ref{related}. 
GeoTrans computes an anomaly score based on the classification of conducted transformations.
Table~\ref{table:mvtec} shows the results for MVTec AD.
Compared to other approaches, our method outperforms existing methods in almost every category, up to a large margin of 15\%.
In all of the 15 categories our method achieves an AUROC of at minimum $84\%$, which shows that our approach is not limited to a specific set of defects or features.
The fact that 1-NN outperforms other competitors except us, demonstrates that our feature extraction and evaluation is well-suited for the problem.

We can observe similar characteristics on MTD, seen in Table~\ref{table:magnets}.
The ROC-Curve in Figure~\ref{fig:magnetroc} shows that our method provides a much higher true positive rate for any false positive rate. DifferNet achieves a recall of about 50\% without any false positive among 191 defect-free test images.
The histogram of anomaly scores is visualized in Figure~\ref{fig:scores}.
There is a large subset of defective samples whose scores differ significantly from all scores of non-defective samples.
The assignment of extremely high scores without any false positive is a characteristic of our method and can be similarly observed for other evaluated product categories.

\begin{table}
\begin{center}
\begin{tabular}{|l|c|}
\hline
Method & AUROC $[\%]$ \\
\hline
GeoTrans\cite{geotrans} & 75.5 \\
GANomaly\cite{ganomaly} & 76.6 \\
DSEBM \cite{dsebm} & 57.2 \\
ADGAN \cite{ADGAN} & 46.4 \\
OCSVM \cite{andrews} & 58.7 \\
1-NN \cite{nazare} & 80.0 \\
\textbf{DifferNet (ours)} & \textbf{97.7} \\
\hline
\end{tabular}
\end{center}
\caption{Area under ROC in \% for detecting anomalies on MTD}
\label{table:magnets}
\end{table}

\subsection{Localization}
\begin{figure*}
\centering
  \includegraphics[width=0.91\textwidth]{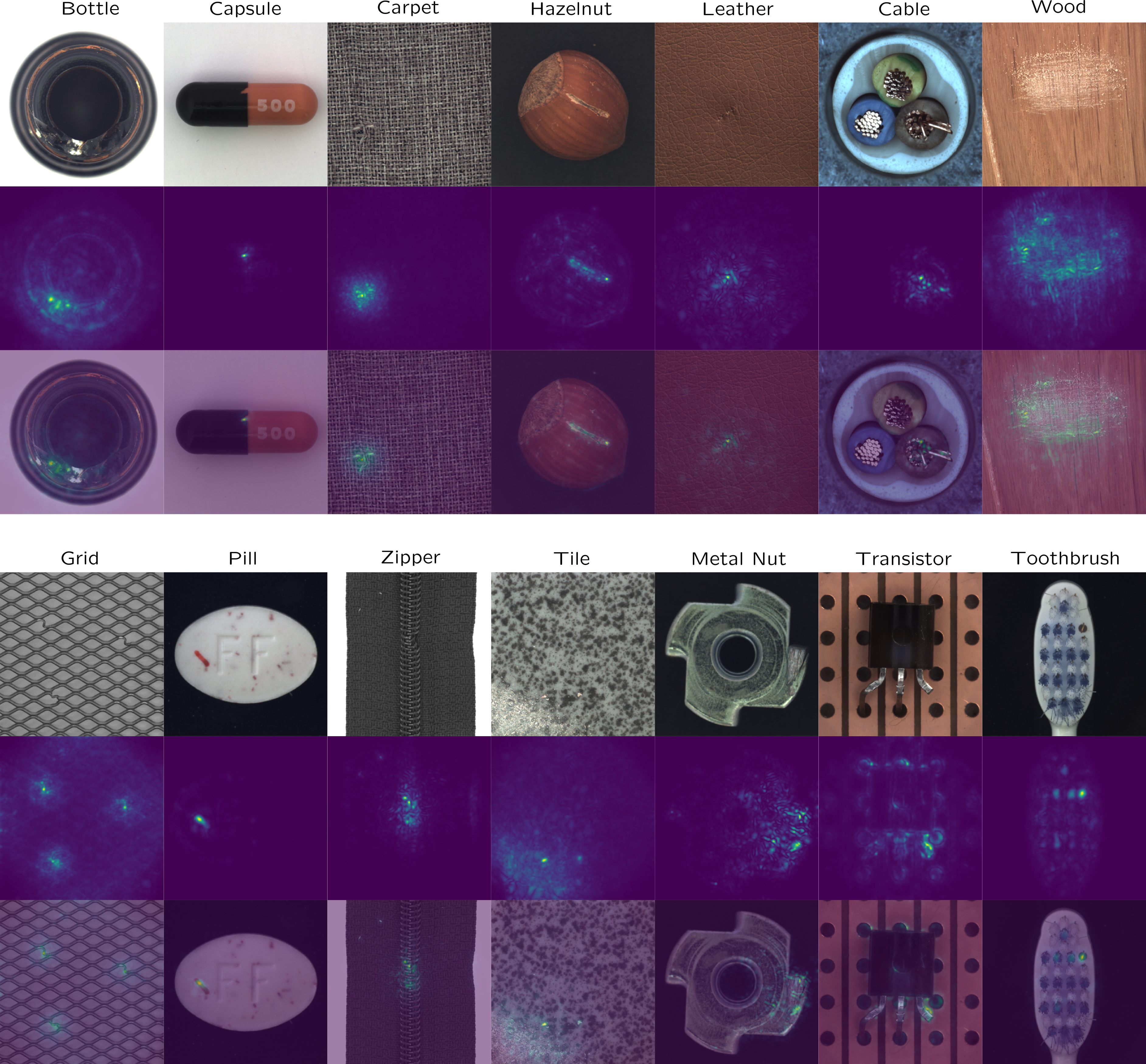} 
 \caption{Localization of anomalous regions of different categories in MVTec AD. The upper rows shows the original anomaly images, the mid rows the localizations provided by DifferNet and the lower rows the superimposition of both. They were generated by backpropagating the negative log-likelihood loss to the input image.}
\label{fig:localization}
\end{figure*}

Results of the localization procedure described in Section~\ref{loc_method} are shown in Figure~\ref{fig:localization}.
The localizations are accurate for many types, sizes and shapes of anomalies; despite the average pooling of feature maps before being processed by the normalizing flow.
Our architecture produces meaningful gradients which can be explained by the models architecture:
First, AlexNet is relatively shallow such that noisy or vanishing gradients are prevented.
Second, the bijectivity of the normalizing flow causes a direct relation between all image features $y$ and all values of $z$ with non-zero gradients.
The gradients tend to appear speckled for larger anomalous regions.
We conject the reason is that pixels, leading to features influencing the anomaly score, are usually not located evenly distributed in the corresponding region.
However, our method enables the human to perceive the defective region and interpret which areas influenced the networks decision to what extent.

\begin{figure}
\centering
  \includegraphics[width=0.45\textwidth]{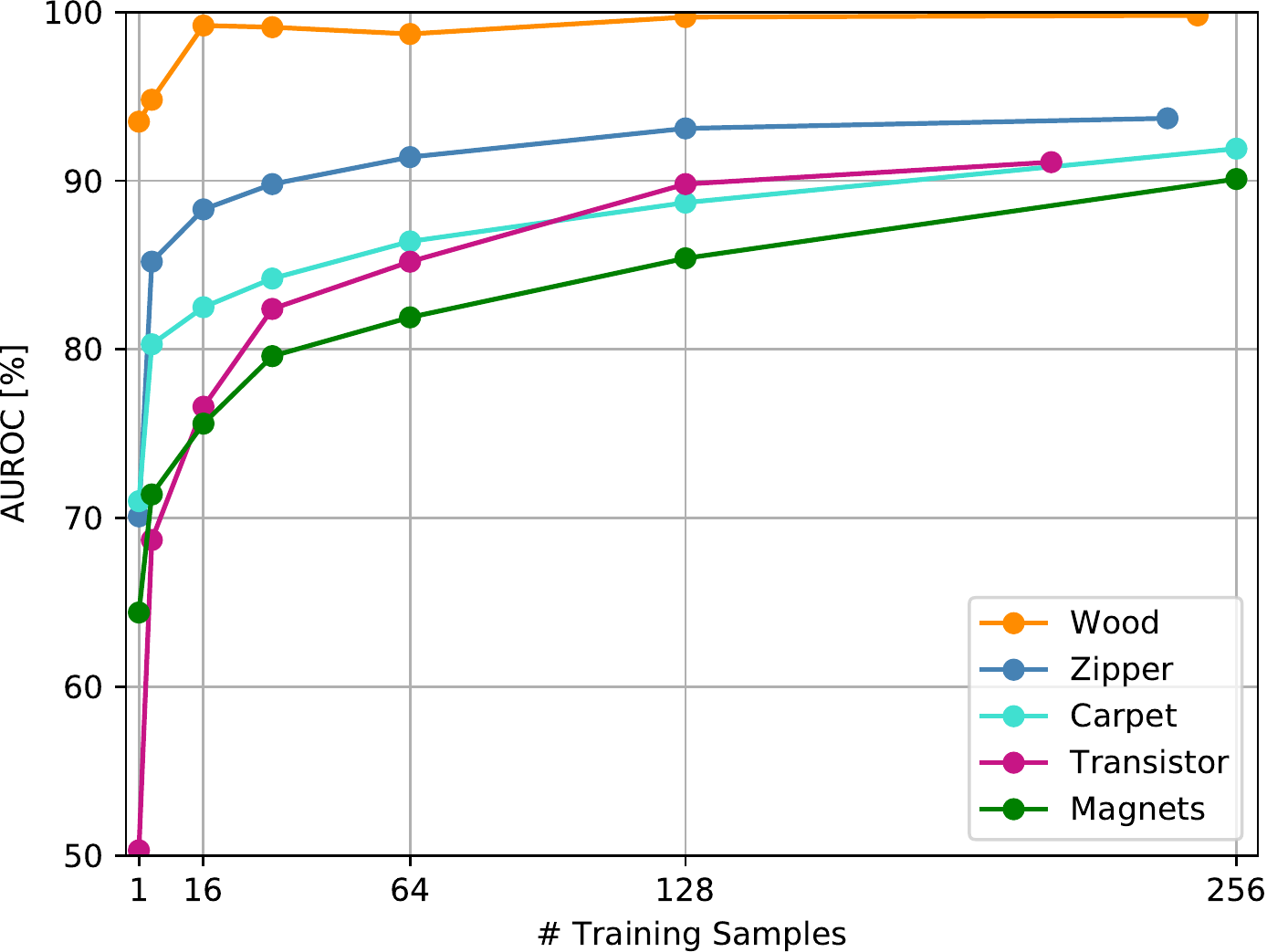} 
 \caption{Detection performance of DifferNet, measured by AUROC, depending on the training set size of MTD and of some categories of MVTec AD. Best viewed in color.
 }
\label{fig:samples}
\end{figure}




\begin{table}
\begin{center}
\begin{tabular}{|c|c|c|c|c|c|c|c|}
\hline
\footnotesize{Config.} & \footnotesize{A} & \footnotesize{B} & \footnotesize{C} & \footnotesize{D} & \footnotesize{E} & \footnotesize{F} & \footnotesize{G}\\
\hline
\footnotesize{multi-scale} & \underline{\xmark}& \underline{\xmark}&\checkmark     & \checkmark  & \checkmark  & \checkmark& \checkmark  \\
\footnotesize{train transf.} & \underline{\xmark}&  \checkmark& \underline{\xmark} & \checkmark & \checkmark &\checkmark & \checkmark  \\
\footnotesize{\# test transf.} &\footnotesize{\underline{1}}&\footnotesize{64}& \footnotesize{\underline{1}} & \footnotesize{\underline{1}} &  \footnotesize{\underline{4}} & \footnotesize{\underline{16}} & \footnotesize{64}   \\
\hline
\footnotesize{AUROC $[\%]$} & \footnotesize{84.4} & \footnotesize{90.2} & \footnotesize{86.6}  & \footnotesize{86.5} & \footnotesize{91.6} & \footnotesize{94.1}  & \footnotesize{\textbf{94.9}}\\ %
\hline
\end{tabular}
\end{center}
\caption{Average detection performance for all categories of MVTec AD when modifying our proposed training and evaluation strategy. The columns show parameter configurations named from A to F. Parameters that differ from our proposed configuration are underlined.}
\label{table:ablation}
\end{table}
\subsection{Ablation Studies}
\label{ablation}
To quantify the effects of individual strategies used in our work, we performed an ablation study by comparing the performance on MVTec AD \cite{mvtec} when modifying the strategies. In addition, the model's behavior for different characteristics of the training set is analyzed.\\
\textbf{Preprocessing Pipeline and Evaluation.}
Table~\ref{table:ablation} compares the detection performance on MVTec AD for different configurations regarding multi-scaling, the usage of transformations in training and the number of used transformations $T_i$ for evaluation.
Having one test transformation means that only the original image was used for evaluation.
Note that we outperform existing methods even without the proposed transformations and multi-scale strategy.
Since relevant features could appear at any scale, it is beneficial to include features at multiple scales which is shown by an AUROC improvement of $4.7\%$.
Having transformed samples in training is crucial as it enables multi-transform evaluation and helps for generalization and data augmentation.
The similar performances of configuration C and D reflect that applying transformations in training is only useful if they are performed in inference as well.
The more of these transformations are then used, the more meaningful the score is, as the rising performance of configurations D to G shows.\\
\textbf{Number of Training Samples.}
We investigate the effect of the training set size on the detection performance as shown on Figure~\ref{fig:samples} and on the right of Table~\ref{table:mvtec}. 
The reported results are the average over three runs with different random subsets per training set size.
It can be seen that our model and training procedure allows for a stable training even on small training sets.
This can be explained by the usage of multiple transformations and the averaging of feature maps.
Our model profits from this strategy which is a mixture of augmentation and compression.
DifferNet requires only 16 training samples to outperform existing approaches that use the whole training set with 369 samples per class on average.
For some classes, the subsets cannot represent the feature variation of normal samples.\\
\textbf{Multimodality.}
The feature distributions of the evaluated categories are unimodal.
We also investigated the performance on multimodal distributions. 
Therefore, we also observed the detection performance when using all 15 categories of MVTec as training data.
To capture this more complex distribution, we used 12 coupling blocks.
The result is a mean AUROC of 90.2\% which shows that our method is able to handle multimodal distributions well.
The regressing sub-blocks inside the NF appear to capture the modes and switch between them depending on their input.

\section{Conclusion}
We presented \textit{DifferNet} to detect defects in images by utilizing a normalizing-flow-based density estimation of image features at multiple scales.
Likelihoods of several transformations of a single image are used to compute a robust anomaly score.
Therefore, there is no need for a large amount of training samples.
The design and scoring function is chosen such that image gradients can be exploited to localize defects. 
As shown, the method also scales to multimodal distributions which resembles real-world settings.
In the future we plan to refine the concept in order to find anomalies in video data comparable to \cite{wentong_ano, wentong_ano2}.

\subsubsection*{Acknowledgements} This work was funded by the Deutsche Forschungsgemeinschaft (DFG, German Research
Foundation) under Germany's Excellence Strategy within the Cluster of Excellence PhoenixD (EXC 2122).

\newpage
\clearpage
{\small
\bibliographystyle{ieee_fullname}
\bibliography{SSBD}
}

\end{document}